\documentclass{article}
\usepackage{ijcai25}

\usepackage{times}
\usepackage{soul}
\usepackage{url}
\usepackage[hidelinks]{hyperref}
\usepackage[utf8]{inputenc}
\usepackage[small]{caption}
\usepackage{graphicx}
\usepackage{amsmath}
\usepackage{amsthm}
\usepackage{booktabs}
\usepackage{algorithm}
\usepackage{algorithmic}
\usepackage[switch]{lineno}

\usepackage{multicol}
\usepackage{multirow}

\usepackage{amssymb}

\usepackage{subfigure}

\title{Adapting Large Language Models for Time Series Modeling via a Novel Parameter-efficient Adaptation Method}

\author{
Juyuan Zhang$^1$
\and
Wei Zhu$^2$\and
Jiechao Gao$^{3}$\\
\affiliations
$^1$Nanyang Technological University, Singapore\\
$^2$University of Hong Kong, Hong Kong, China\\
$^3$University of Virginia, VA, United States\\
\emails
JZHANG161@e.ntu.edu.sg,
michaelwzhu91@gmail.com,
jg5ycn@virginia.edu
}

\begin{document}

\maketitle

\begin{abstract}

Time series modeling holds significant importance in many real-world applications and has been extensively studied. While pre-trained foundation models have made impressive strides in the fields of natural language processing (NLP) and computer vision (CV), their development in time series domains has been constrained by data sparsity. A series of recent studies have demonstrated that large language models (LLMs) possess robust pattern recognition and reasoning abilities over complex sequences of tokens. However, the current literature have yet striked a high-quality balance between (a) effectively aligning the time series and natural language modalities, and (b) keeping the inference efficiency. To address the above issues, we now propose the Time-LlaMA framework. Time-LlaMA first converts the time series input into token embeddings through a linear tokenization mechanism. Second, the time series token embeddings are aligned with the text prompts. Third, to further adapt the LLM backbone for time series modeling, we have developed a dynamic low-rank adaptation technique (D-LoRA). D-LoRA dynamically chooses the most suitable LoRA modules at each layer of the Transformer backbone for each time series input, enhancing the model's predictive capabilities. Our experimental results on an extensive collection of challenging real-world time series tasks confirm that our proposed method achieves the state-of-the-art (SOTA) performance.\footnote{Codes will be made public upon acceptance. }

\end{abstract}

\section{Introduction}

Time series forecasting (TSP) represents a crucial modeling endeavor \cite{jin2023large}, spanning a wide array of practical applications such as climate modeling, inventory management, and energy demand prediction. Typically, each forecasting task demands specialized domain expertise and bespoke model architectures. This requirement has precluded the development of a robust foundational model (FM) capable of few-shot or zero-shot learning, akin to GPT-3 \cite{brown2020language}, GPT-4 \cite{gpt4}, and Claude-3\footnote{https://claude.ai/}, within the time series domain. Despite the fact that time series modeling has yet to witness similar groundbreaking advancements, the remarkable capabilities of large language models (LLMs) have fueled interest in their application to time series forecasting tasks \cite{zhou2023one}.

Despite the advancements in the literature on Large Language Model (LLM)-based Time Series (TS) modeling \cite{zhou2023one,jin2023time}, several limitations remain. Firstly, the successful integration of time series data with natural language in LLM-based TS modeling depends heavily on the appropriate alignment of their respective modalities. Current approaches primarily rely on text prompts and cross-attention mechanisms, which do not effectively leverage the vocabulary. Secondly, recent studies adopt a methodology similar to PatchTST \cite{nie2022time}, transforming a univariate time series into a sequence of patches that are then treated as tokens input into Transformer blocks. This approach necessitates converting multivariate Time Series Prediction (TSP) tasks into multiple univariate TSP subtasks, leading to increased inference latency. Lastly, the current works maintains the LLM backbone in a frozen state and refrains from incorporating additional trainable components within the Transformer blocks \cite{jin2023time}, which may limit the models' ability to adapt to specific tasks more effectively.

To address the above issues, we introduce Time-LlaMA, an innovative framework designed to harness large language models for time series forecasting. Our approach diverges from prior methodologies \cite{zhou2023one,jin2023time} in the following aspects. First, we treat each channel within multivariate time series data as an individual token. These tokens are processed through a straightforward multi-layer perceptron (MLP) before they are input into the Transformer architecture. Furthermore, we employ a trainable cross-attention module to align the tokenized time series data with the embeddings of the text prompt, rather than the entire vocabulary, thereby enhancing the model's focus on relevant information. Notably, the text prompt is not passed through the Transformer backbone to minimize inference delay. Additionally, we present D-LoRA, a novel variant of the LoRA technique that incorporates a mixture-of-experts mechanism. D-LoRA dynamically assigns distinct sets of LoRA modules to various input samples, leading to improved performance across the board. Extensive experimentation has proved that our Time-LlaMA method surpasses recent state-of-the-art baseline methods. The contributions of our work are summarized as follows:
\begin{itemize}
\item We propose a novel framework Time-LlaMA, which successfully incorporate multi-variate TS modeling with LLMs. By aligning to text prompts and fine-tuning the LLMs with a novel D-LoRA method, our work pushs the limit of LLM based TS modeling methods. 

\item Time-LlaMA consistently exceeds state-of-the-art performance in TS forecasting tasks, especially in few-shot and zero-shot scenarios. Moreover, this superior performance is achieved while maintaining excellent parameter efficiency and inference efficiency. 

\end{itemize}

\begin{figure*}[t]
\centering
\includegraphics[width=0.8\textwidth]{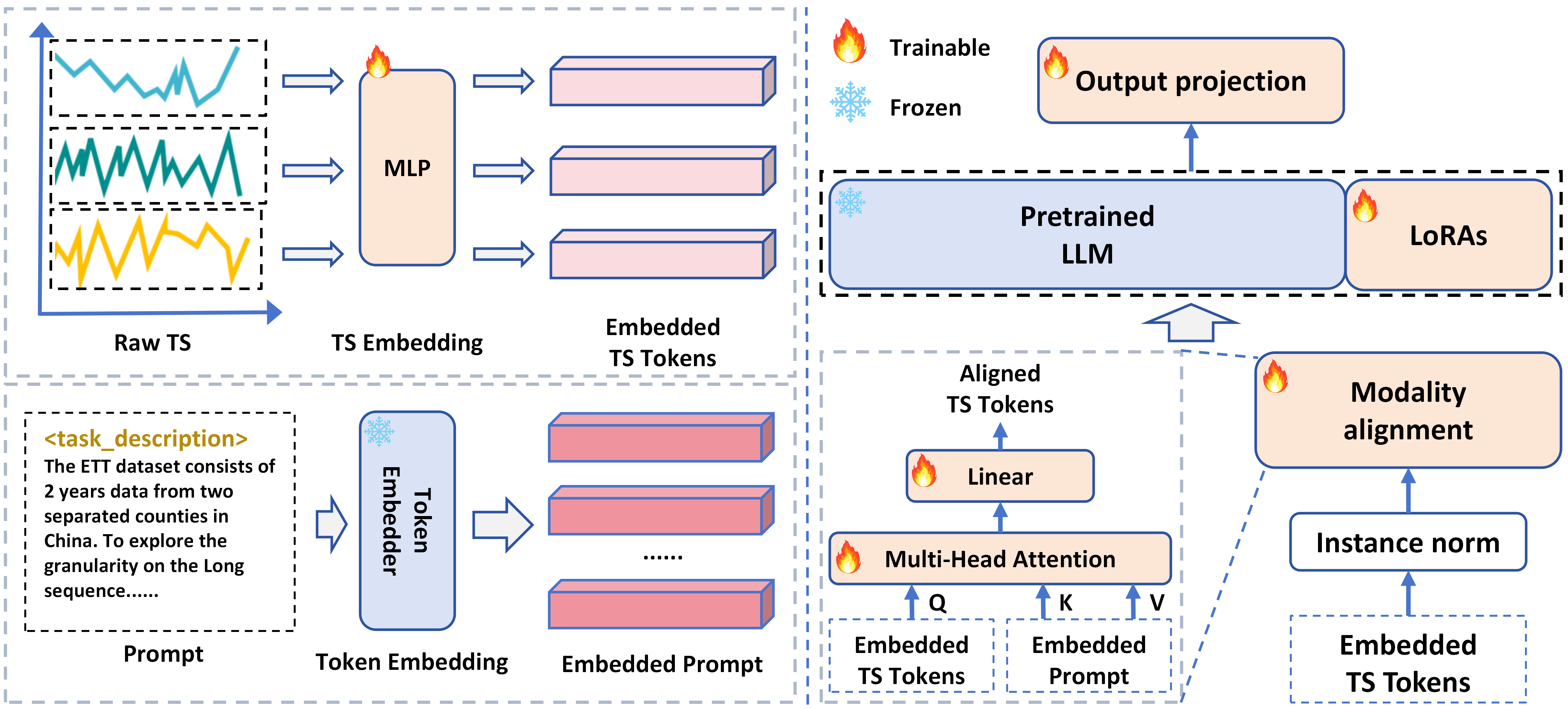} 
\caption{Schematic illustration of our Time-LlaMA framework.}
\label{fig:framework}
\end{figure*}

\begin{figure*}[t]
\centering
\includegraphics[width=0.62\textwidth]{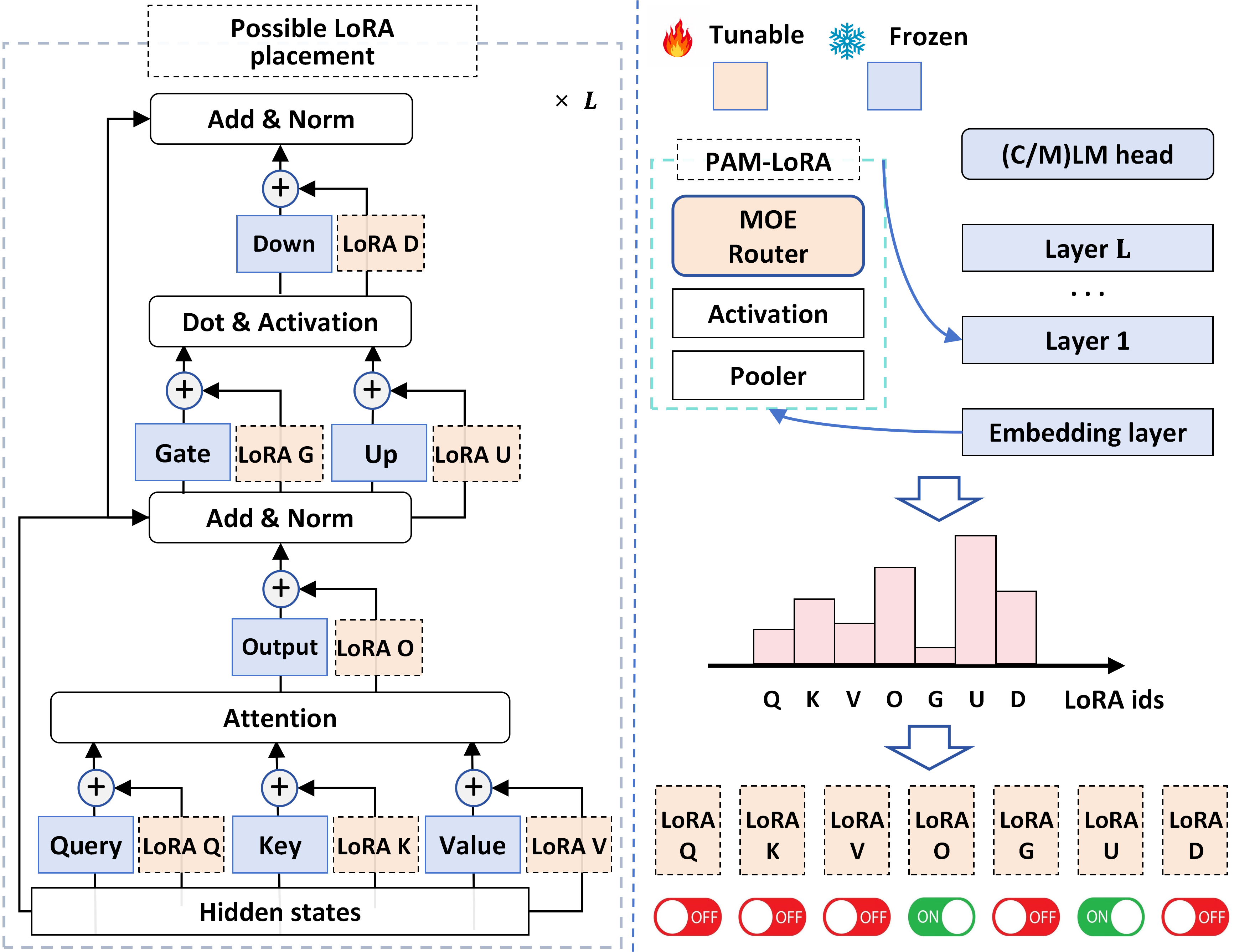} 
\caption{Schematic illustration of our D-LoRA method.  }
\label{fig:dyna_lora_architecture}
\end{figure*}

\section{Related Work}

\noindent\textbf{Time series modeling.} \quad The progressive advancements in natural language processing and computer vision have led to the development of sophisticated Transformer \cite{Vaswani2017AttentionIA} variants tailored for a wide array of time series forecasting applications \cite{zhou2021informer,wu2021autoformer}. Central to these innovations is the methodology by which Transformers handle time series data. For instance, I-Transformer \cite{liu2023itransformer} treats each univariate time series as a distinct token, forming multivariate time series into sequences of such tokens. More recently, PatchTST \cite{nie2022time} adopts an assumption of channel independence, transforming a univariate time series into multiple patches, which are subsequently treated as tokens and processed through a Transformer encoder. This approach has yielded notable results on various benchmark datasets for time series. Nevertheless, these forecasting models are trained end-to-end using task-specific datasets. A recent trend involves the developments of Transformer-based foundational models for time series analysis \cite{das2023decoder,goswami2024moment} via pre-training, capable of being swiftly adapted to diverse downstream tasks.

\noindent\textbf{Cross-modal transfer learning using language models} \quad Recent investigations have highlighted the efficacy of transferring Transformer models \cite{Vaswani2017AttentionIA}, which are pretrained on extensive textual corpora, to other modalities. \cite{lu2022frozen} employs a frozen pretrained Transformer across a spectrum of sequence classification tasks encompassing numerical computation, vision, and protein structure prediction, training only the newly introduced classification heads. ORCA \cite{shen2023cross} adopts an align-then-refine workflow to adapt to target tasks. Specifically, given the target input, ORCA initially learns an embedding network that aligns the feature distribution of the embedded data with that of the pretraining modality. Subsequently, the pretrained model is fine-tuned on the aligned data to harness cross-modal knowledge. Building upon these capabilities, recent studies have successfully adapted large language models (LLMs) for time series analysis through the use of a reprogramming module and a tokenization technique, while maintaining the LLMs in a frozen state \cite{zhou2023one,jin2023time}. Our contribution to this body of research is twofold: (a) we conceptualize each time series variable as a token, enabling simultaneous predictions for all variables within a single forward pass, thereby enhancing efficiency. (b) We introduce a novel LoRA methodology that fine-tunes the LLM backbone in a parameter-efficient manner, advancing the state-of-the-art in LLM-based time series modeling.

\noindent\textbf{Parameter efficient fine-tuning for pretrained Transformer models} \quad Parameter-efficient fine-tuning (PEFT) optimizes a small portion of added parameters when fine-tuning a LLM and keeps the backbone model frozen \cite{Ding2022DeltaTA,Zhang2023LearnedAA}. LoRA \cite{hu2021lora} is inspired by \cite{aghajanyan-etal-2021-intrinsic} and \cite{2018arXiv180408838L}, and hypothesizes that the change of weights during model fine-tuning has a low intrinsic rank and optimizes the low-rank decomposition for the change of original weight matrices. LoRA \cite{hu2021lora} is proven to be effective and yield stable results when applied to both relatively small pretrained backbones and large language models \cite{2023arXiv230514314D,PromptCBLUE}. However, the original LoRA paper does not specify how to add LoRA modules of different ranks to the Transformer backbones for adapting different tasks. In this work, we propose a novel LoRA variant that can help the LLM backbone to better adapt to the time series prediction tasks and achieve state-of-the-art performance.

\section{Methodology}

Our model architecture is illustrated in Figure \ref{fig:framework}. We concentrate on reprogramming a language foundation model, such as Llama-2 \cite{Touvron2023Llama2O}, for general time series forecasting, by introducing additional parameters that constitute less than 1.0\% of the backbone model's total parameters. This work targets both univariate and multivariate time-series prediction tasks. Our approach comprises three primary components: (a) the time-series tokenization module, which includes a linear projection layer that transforms the time-series input into a sequence of token embeddings. (b) The modality alignment module, which ensures that the time-series inputs are aligned with the embeddings of the prompts. (c) The pre-trained LLM backbone, which is fine-tuned using a novel dynamic LoRA (D-LoRA) technique. Finally, the output projection module generates the prediction results.

\subsection{Preliminaries}

\noindent \textbf{Transformer model} \quad As depicted in Figure \ref{fig:dyna_lora_architecture}, each Transformer layer of a LLM with $L$ layers such as LlaMA-2 \cite{Touvron2023Llama2O} consists of a multi-head self-attention (MHA) module and a fully connected feed-forward (FFN) sub-layer. MHA contains four linear modules, which are the Query (Q), Key (K), Value (V), and Output (O) modules. FFN contains three linear modules: Gate (G), Up (U), and Down (D). For notation convenience, we will refer to the number of modules in a Transformer block as $N_{mod}$. Thus, in LlaMA-2, $N_{mod} = 7$.

\noindent \textbf{LoRA} \quad For any Transformer module $m \in \text{\{Q, K, V, O, G, U, D\}}$, the LoRA method adds a pair of low-rank matrices to reparameterize its weights. Formally, the forward calculation of module $m$ in layer $l$ with LoRA is: 
\begin{equation}
x^{'} = xW_{m, l} + g_{m, l} * xW_{m, l}^{A}W_{m, l}^{B} + b_{m, l}, 
\label{eq:lora_eq_1}
\end{equation}
where $W_{m, l} \in \mathbf{R}^{d_1 \times d_2 }$ is the weight matrix of module $m$, $b_{m, l}$ is its bias term. $W_{m, l}^{A} \in \mathbb{R}^{d_{1} \times r }$ and $W_{m, l}^{B} \in \mathbb{R}^{ r \times d_{2} }$ are the low-rank matrices for the LoRA module, and $r \ll \min (d_{1}, d_{2})$. $r$ is the rank of the two matrices and will also be referred to as the rank of the LoRA module. Here, we include a binary gate $g_{m, l} \in \{0,  1\}$ to conveniently control the inclusion of LoRA $m$ in the forward calculation. For the vanilla LoRA method, all the LoRA gates $g_{m, l}$ are set to 1.

\subsection{Task formulation}  

In this study, we address the challenge of multivariate time series prediction. Given a sequence of historical observations $\mathbf{X} \in \mathcal{R}^{N \times T_L}$ consisting of $N$ different 1-dimensional variables across $T_L$ time steps, we aim to adapt a large language model $f(\cdot)$ to understand the input time series and  accurately forecast the values at $T_P$ future time steps, denoted by $\mathbf{Y} \in \mathcal{R}^{N \times T_P}$.

\subsection{Token Embedding} 


The recent work like Time-LLM only supports the uni-variate time series with the patching mechanism, resulting in inconvenience when dealing with multi-variate time series data. Thus, in order to seamlessly apply the LLM to time series prediction, we consider the $i$-th variate $X_{i, :}$'s whole series as a token, and embedd it with:
\begin{equation}
\mathbf{h}_{i}^{TS, 0} = \text{TSEmb}(X_{i, :}),
\end{equation}
where $\text{TSEmb} \text{ : } \mathcal{R}^{T} \mapsto  \mathcal{R}^{d_{m} } $ denotes the time-series token embedding module, $d_{m}$ denotes the hidden dimension of the LLM backbone. $\text{TSEmb}$ is implemented by multi-layer perceptron (MLP).  The obtained tokens of time-series variates interact with one another by self-attention and are independently processed by the shared feed-forward network in the LLM backbone. And $\mathbf{H}^{TS, 0} = \{ \mathbf{h}_{1}^{TS, 0}, ..., \mathbf{h}_{N}^{TS, 0} \}$ denotes the whole token sequences of the input time series.

\subsection{Alignment}

Note that time series is different from the language modality, making it difficult for the LLM to understanding time series. To close this gap, we propose to align the time-series token embeddings $\mathbf{H}^{0}$ with the prompts' embeddings $\mathbf{H}^{P, 0}$. To realize this alignment, we utilize a multi-head cross-attention (MHCA) layer where $\mathbf{H}^{0}$ acts as the query tensor and $\mathbf{H}^{P, 0}$ acts as the key and value tensor. Specifically, for each attention head $k \in \{1, 2, ..., K\}$, we define the query tensors as $Q_{k} = \mathbf{H}^{0} W_{k}^{Q}$, the key tensors as $K_{k} = \mathbf{H}^{P, 0} W_{k}^{K}$, and the value tensors as $V_{k} = \mathbf{H}^{P, 0} W_{k}^{V}$, where $W_{k}^{Q}, W_{k}^{K}, W_{k}^{v} \in \mathcal{R}^{ d_{m} \times d_{head} } $ are the weight matrices, $d_{head} = d_{m} / K$ is the hidden dimension on each head. Then the time-series token embeddings are aligned to the natural language representation via the following equations:
\begin{equation}
\begin{aligned}
A_{k} = &  \text{Softmax}( \dfrac{Q_{k} K_{k}^{\intercal} }{ \sqrt{d_{head}} } )   \\ 
\mathbf{H}^{0} \leftarrow &  \mathbf{H}^{0} + \text{Concat}([A_{1}, ..., A_{K}])  W^{O},\\
\end{aligned}
\label{eq:modality_align}
\end{equation}
where $\text{Concat}()$ is the concatenation operation, and $W^{O} \in \mathcal{R}^{ d_{m} \times d_{m} } $ is the attention output projection matrix. Then the input for the LLM's Transformer blocks $\mathbf{H}^{0}$ is obtained by projecting $\mathbf{H}^{0}$ to dimension $d_{model}$, the hidden dimension of the LLM.

\subsection{D-LoRA} 

In the previous works \cite{zhou2023one,jin2023time} on applying LLM backbones to the time series tasks, the LLMs are kept entirely frozen, making it convenient for task adaptation. However, this setting restricts the expressiveness of the whole model. Inspired by the recent works on parameter-efficient fine-tuning in the LLM research, we propose to fine-tune the LLM backbone in a parameter-efficient manner when adapting it to time-series tasks. However, as our initial experiments indicate, the vanilla LoRA method \cite{hu2021lora} does not perform well on time-series prediction tasks. And if we change the LoRA settings on which Transformer linear module are modified by a LoRA module, this method's performance would improve. We hypothesize that when adapted to the time-series tasks, how LLM process the data is different from encoding the pure text data, thus how to set the LoRA modules should be different. In this work, we take a step further and propose an input-adaptive dynamic LoRA (D-LoRA) method, which dynamically assign LoRA modules to the different Transformer modules based on the input.

We now present the details of our D-LoRA method. The core of D-LoRA is the input-dependent LoRA assignment mechanism, as shown in Figure \ref{fig:dyna_lora_architecture}. Under this mechanism, a LoRA router takes the input's hidden states as input and outputs the assigned LoRA experts for the current layer. Denote the hidden state of the input right before the Transformer layer $l$ as $\mathbf{H}^{l-1} \in \mathbf{R}^{N \times d_{m} }$. Then a pooling operation transforms it to a single vector $\mathbf{h}_{pooled}^{l} \in \mathbf{R}^{1 \times d_{m} }$:
\begin{equation}
\mathbf{h}_{pooled}^{l} = \text{Pooler}( \mathbf{H}^{l-1} ).
\end{equation}
Consistent with \cite{radford2018improving} and \cite{lewis2019bart}, $\text{Pooler}()$ takes the vector representation of the last token in the input as $\mathbf{h}_{pooled}^{l}$. Then, $\mathbf{h}_{pooled}^{l}$ will go through an activation function $g$ and then the LoRA router $R^{l}$ right before layer $l$. $R^{l}$ assigns the current input to the most suitable LoRA modules. This router contains (a) a linear layer that computes the probability of $\mathbf{h}^{l}$ being routed to each LoRA module $\text{LoRA}_{m}$ ($m \in $ \{Q, K, V, O, G, U, D\}), (b) a softmax function to model a probability distribution over the LoRA modules, and finally, (c) a $\text{Top\_K}(\cdot, n)$ function that choose the top $n > 0$ experts with the highest probability masses. Formally, 
\begin{equation}
R^{l}(\mathbf{h}^{l}) = \text{Top\_K}( \text{Softmax} ( g(\mathbf{h}^{l}) W_{r}^{l} ), n),
\label{eq:router}
\end{equation}

where $W_{r}^{l} \in \mathbf{R}^{d_{m} \times N_{mod}}$ is the router's weight. $R^{l}(\mathbf{h}^{l})$ is a $N_{mod}$-dim vector, in which the $m$-th element is a binary value in \{0, 1\} and is assigned to $g_{m, l}$ to activate or deactivate LoRA $m$:
\begin{equation}
g_{m, l} \leftarrow R^{l}(\mathbf{h}^{l})[m],
\end{equation}
and $\sum_{m = 1}^{N_{mod}} g_{m, l}$ equals $n$. The LoRA router dynamically selects and activates the best $n > 0$ experts for each input during inference.

Different from the standard LoRA method \cite{hu2021lora}, our work: (a) determines the assigned LoRA modules at the Transformer's layer level, selecting which Transformer module should be modified by its corresponding LoRA module. (b) The decision on selecting LoRA modules are conditioned on the input data, and different test samples could set LoRA modules differently. (c) Note that for a test input, different Transformer layers may choose to assign different LoRA modules. (d) Note that we can adjust the number of assigned LoRA modules $n$ per layer, making inference more efficient.

\subsection{Output layer}

After $\mathbf{H}^{0}$ is encoded by the LLM, we obtain the output representation $\mathbf{H}^{L}$. Then $\mathbf{H}^{L}$ will go through a linear layer to obtain the predictions for the future $T_{P}$ time steps:
\begin{equation}
\hat{ \mathbf{Y} } = \mathbf{H}^{L} W^{P} + b^{P},
\end{equation}
where $W^{P} \in \mathcal{R}^{ d_{m} \times T_{P} } $ is the weight matrix, and $b^{P} \in \mathcal{R}^{ 1 \times T_{P} } $ is the bias term.

\subsection{Loss calculations}

Following the standard practice for the time-series prediction tasks, the objective is to  minimize the mean square errors between the ground truths $\mathbf{Y}$ and predictions$\hat{ \mathbf{Y} }$:
\begin{equation}
\mathcal{L}_{mse} = \| \mathbf{Y} - \hat{ \mathbf{Y} } \|_{F}^{2}. 
\end{equation}

Following \cite{fedus2022switch}, we add a load balancing loss to the training loss function. Consider a training batch $B$ with $N_{B}$ samples, let $f_{i}^{l}$ represent the proportion of prompts assigned to the $i$-th LoRA expert in layer $l$, 
\begin{equation}
f_{i}^{l} = \dfrac{1}{N_{B}} \sum_{x\in B} \mathbf{1} \{ \arg\max_{j} p^{l}_{j}(x) = i \},
\end{equation}
where $p^{l}_{j}$ is the probability of expert $j$, output by the router $l$. Let $\hat{p}^{l}_{i}$ be the average of probability masses received by the $i$-th expert, $\hat{p}^{l}_{i} = \dfrac{1}{N_{B}} \sum_{x\in B} p^{l}_{i}(x) $. Then, the load balancing loss is given by:
\begin{equation}
\mathcal{L}_{lb} = N_{mod} \sum_{l=1}^{ L } \sum_{i=1}^{ N_{mod} } f_{i}^{l} \cdot \hat{p}^{l}_{i}.
\end{equation}
The $\mathcal{L}_{lb}$ loss term is added to the cross entropy loss with a coefficient $\lambda_{lb} \geq 0$:
\begin{equation}
\mathcal{L} = \mathcal{L}_{mse} + \lambda_{lb} * \mathcal{L}_{lb}.
\end{equation}

\section{Experiments}

\subsection{Baselines} 

We compare our Time-LlaMA method with the SOTA time series models: (a) Time-LLM \cite{jin2023time}, (b) GPT4TS \cite{zhou2023one}, (c) PatchTST \cite{nie2022time}, (d) DLinear \cite{zeng2023transformers}, and (e) TimesNet \cite{wu2022timesnet}.

\subsection{Datasets and evaluation metrics}

For long-term time series forecasting, we assess our Time-LlaMA framework on the following datasets, in accordance with \cite{wu2022timesnet}: ETTh1, ETTm1, Weather, ECL, and Traffic. The evaluation metrics utilized are the mean square error (MSE) and the mean absolute error (MAE). For short-term time series forecasting, we employ the M4 benchmark \cite{makridakis2018m4}. The evaluation metrics for this benchmark include the symmetric mean absolute percentage error (SMAPE), the mean scaled absolute error (MSAE), and the overall weighted average (OWA). The detailed introductions to the datasets can be found in the Appendix.

\subsection{Experimental setups}

We use Llama-2 7B \cite{Touvron2023Llama2O} as the default backbone unless stated otherwise, thus $d_{m} = 4096$. We utilize the first $L = 8$ Transformer blocks of LlaMA-2 7B for our Time-LlaMA framework. For the alignment module, the number of attention heads is $K=8$. For D-LoRA, the LoRA rank is set to $r = 4$, and each layer will select $n=4$ LoRA modules during inference. More details for the training hyper-parameters are presented in the Appendix.

\subsection{Main results}

\noindent \textbf{Results for long-term forecasting} \quad For the long-term forecasting tasks, the input time series length $T_{L}$ is set as 512, and we use four different prediction horizons $T_{P} \in \{96, 192, 336, 720\}$ ($H \in \{24, 36, 48, 60\}$ for the ILI task). The evaluation metrics include mean square error (MSE) and mean absolute error (MAE). In Table \ref{tab:results_main_long_term}, we report the average score over four different horizons.

The experimental results demonstrate that our Time-LlaMA method outperforms the baselines on most of the (task, prediction horizon) pairs. The comparison against Time-LLM \cite{jin2023time} and GPT4TS \cite{zhou2023one} is particularly meaningful. These two are very recent works on adapting large language models to the time-series forecasting tasks. When compared to the pre
vious state-of-the-art (SOTA) model PatchTST which is trained from scratch on each task, Time-LlaMA can also achieves advantages.

\begin{table*}
\centering
\resizebox{0.86\textwidth}{!}{
\renewcommand\arraystretch{1.12}
\begin{tabular}{cc|cc|cc|cc|cc|cc|cc}
\hline
\multicolumn{2}{c}{\textbf{Methods}}   &   \multicolumn{2}{c}{\textbf{Time-LlaMA}}    &   \multicolumn{2}{c}{\textbf{TIME-LLM}}     &     \multicolumn{2}{c}{\textbf{GPT4TS}}     &     \multicolumn{2}{c}{\textbf{PatchTST}}   &   \multicolumn{2}{c}{\textbf{DLinear}}    &   \multicolumn{2}{c}{\textbf{TimesNet}}    \\ 

\multicolumn{2}{c}{\textbf{Metric}}   &  MSE   &  MAE    &  MSE   &  MAE     &  MSE   &  MAE     &  MSE   &  MAE     &  MSE   &  MAE    &  MSE   &  MAE       \\
\hline

\multirow{4}*{ETTh1}   &   96    &    0.377  &    0.398     &      0.386    &   0.409     &    \underline{0.376}    &    0.397     &    0.378    &    0.405     &    \textbf{0.375}    &    0.399     &    0.384     &     0.402     \\

  &    192    &    \underline{0.410}  &   0.426     &     0.414    &    0.421     &  0.416     &  0.418     &  0.413     &  0.421     &  \textbf{0.405}     &  0.416     &  0.436     &   0.429  \\
  
&    336      &    \underline{0.421}    &   0.437      &      0.423    &   0.436     &   0.442      &    0.433    &    \textbf{0.422}     &   0.436     &   0.439     &   0.443     &   0.491     &   0.469     \\

&    720      &    \textbf{0.443}    &    0.464     &      0.481    &    0.478      &     \underline{0.477}    &   0.456    &   0.447    &   0.466    &   0.472    &   0.490    &   0.521    &   0.500   \\

\hline



\multirow{4}*{ETTm1}   &   96    &    \underline{0.291}   &   0.343     &       0.298    &    0.356    &     0.292    &   0.346    &   \textbf{0.290}    &   0.342    &   0.299    &   0.343    &   0.338    &   0.375     \\

  &    192    &    \textbf{0.326}    &   0.366     &       0.334    &   0.377     &    \underline{0.332}    &   0.372    &   0.332    &   0.369    &   0.335    &   0.365    &   0.374    &   0.387    \\
  
&    336      &    \textbf{0.352}    &    0.384     &      \underline{0.365}    &   0.389     &   0.366   &   0.394   &   0.366   &   0.392   &   0.369   &   0.386   &   0.410   &   0.411  \\

&    720      &    \textbf{0.405}    &    0.416     &       \underline{0.413}    &    0.418     &   0.417   &   0.421   &   0.416   &   0.420   &   0.425   &   0.421   &   0.478   &   0.450  \\

\hline 



\multirow{4}*{Weather}   &   96    &   \underline{0.151}  &   0.207  &  0.154   &  0.208   &    0.162  &    0.212   &   \textbf{0.149}    &   0.198    &   0.176    &    0.237    &  0.172   &  0.220   \\
 
  &    192    &    \textbf{0.193}    &   0.240     &    0.198    &     0.247     &   0.204    &    0.248     &    \underline{0.194}    &    0.241   &    0.220   &   0.282    &   0.219    &   0.261  \\
  
&    336      &    \textbf{0.242}    &   0.287     &    0.251    &   0.282     &   0.254   &   0.286   &   \underline{0.245}  &     0.282   &   0.265   &   0.319   &    0.280   &    0.306   \\

&    720      &    \textbf{0.313}    &   0.332    &    0.317  &  0.338    &   0.326   &   0.337    &   \underline{0.314}    &  0.334    &  0.333    &  0.362    &  0.365    &   0.359  \\

\hline

\multirow{4}*{ECL}   &   96    &    \textbf{0.128}    &   0.224     &      0.137    &    0.235     &    0.139     &    0.238     &    \underline{0.129}     &    0.222     &    0.140     &    0.237     &    0.168     &    0.272    \\

  &    192    &    \textbf{0.152}    &    0.247     &       0.158    &   0.242     &   \underline{0.153}     &    0.251     &    0.157     &    0.240     &    0.153     &    0.249     &    0.184     &    0.289  \\
  
&    336      &    \textbf{0.161}    &   0.256     &       0.164    &   0.261     &   0.169     &    0.266     &    \underline{0.163}     &    0.259     &    0.169     &    0.267     &    0.198     &    0.300  \\

&    720      &    \underline{0.198}     &    0.292     &       0.204    &    0.293     &   0.206     &    0.297     &    \textbf{0.197}     &    0.290     &    0.203     &    0.301     &    0.220     &    0.320   \\

\hline

\multirow{4}*{Traffic}   &   96    &    \underline{0.379}  &  0.270     &      0.382    &   0.274     &   0.388     &    0.282     &    \textbf{0.378}     &    0.269     &    0.410     &    0.282     &    0.593     &    0.321     \\

  &    192    &    \textbf{0.396}    &    0.279     &        0.404    &   0.285     &   0.407     &    0.290     &    \underline{0.398}     &    0.280     &    0.423     &    0.287     &    0.617     &    0.336   \\
  
&    336      &    \textbf{0.404}    &   0.282     &       0.410    &   0.291     &   0.412     &    0.294     &    \underline{0.406}    &    0.282     &    0.436     &    0.296     &    0.629     &    0.336   \\

&    720      &    \textbf{0.446}   &    0.306     &       0.456    &   0.308     &   0.450     &    0.312     &    \underline{0.448}     &    0.307     &    0.466     &    0.315     &    0.640     &    0.350  \\



\hline 
\end{tabular}}

\caption{\label{tab:results_main_long_term} Results for the long-term forecasting tasks. The prediction horizon $T_{P}$ is one of $\{24, 36, 48, 60\}$ for ILI and one of \{96, 192, 336, 720\} for the others. Lower value indicates better performance. \textbf{Bold} values represent the best MSE score, while \underline{Underlined} means the second best MSE score.} 

\end{table*}

\begin{table*}[tb!]
\centering
\resizebox{0.66\textwidth}{!}{
\renewcommand\arraystretch{1.2}
\begin{tabular}{c|ccccccccccccccc}
\hline
\textbf{Methods}   &   \textbf{Time-LlaMA}    &   \textbf{TIME-LLM}     &     \textbf{GPT4TS}     &     \textbf{PatchTST}   &   \textbf{DLinear}    &   \textbf{TimesNet}    \\ 

\hline

\emph{SMAPE}   &   \textbf{11.96}  &  \underline{12.01}   &   12.69   &    12.06      &    13.63  &  12.88       \\

\emph{MSAE}   &    \textbf{1.656}  &   \underline{1.663}   &   1.808   &  1.683   &     2.095   &    1.836    \\

\emph{OWA}   &    \textbf{0.881}   &   \underline{0.896}   &   0.942   &    0.905   &   1.051    &    0.955      \\

\hline

\hline
\end{tabular}}

\caption{\label{tab:results_main_short_term} Results for the short-term time series forecasting task, M4. The forecasting horizons are in \{6, 48\}. Lower value indicates better performance. \textbf{Bold} values represent the best score, while \underline{Underlined} means the second best. } 
\end{table*}

\noindent \textbf{Results for short-term forecasting} \quad To demonstrate that our method works in the short-term forecasting tasks, we utilize the M4 benchmark \cite{makridakis2018m4}. Table \ref{tab:results_main_short_term} reports the  SMAPE, MSAE and OWA scores. Our experimental results demonstrate that our Time-LlaMA method consistently surpasses all baselines when conducting short-term time series predictions.

\noindent \textbf{Results for few-shot forecasting} \quad Note that a great property of large language models is its great few-shot learning capability. And it is interesting to investigate whether this capability still stands when they are adapted to model time series. We experiment on the scenarios in which limited training data are available for training, that is, only 5\% of the training time steps in the original training set are utilized for training. We experiment with the Weather and ETTh1 tasks, and the results are presented in Table \ref{tab:results_few_shot}.

From Table \ref{tab:results_few_shot}, we can observe that Time-LlaMA excels over all the strong baseline methods. The comparison between Time-LlaMA and non-LLM method like PatchTST, DLinear or TimesNet demonstrates the advantage of utilizing a pre-trained large language model. The pre-trained LLM contains rich world and semantically knowledge, thus providing a high-quality model parameter initialization for the time-series models. The results underscore the prowess of LLMs as a powerful time series model. The comparison against Time-LLM and GPT4TS emphasize our method's advantage in both knowledge activation and task adaptation, which are directly due to the input-adaptive D-LoRA module and the modality alignment module.

\begin{table*}[tb!]
\centering
\resizebox{0.62\textwidth}{!}{
\renewcommand\arraystretch{1.1}
\begin{tabular}{cc|cc|cc|cc|cc}
\hline
\multicolumn{2}{c}{\textbf{Methods}}   &   \multicolumn{2}{c}{\textbf{Time-LlaMA}}    &   \multicolumn{2}{c}{\textbf{TIME-LLM}}     &     \multicolumn{2}{c}{\textbf{GPT4TS}}     &     \multicolumn{2}{c}{\textbf{PatchTST}}    \\ 

\multicolumn{2}{c}{\textbf{Metric}}   &  MSE   &  MAE    &  MSE   &  MAE     &  MSE   &  MAE     &  MSE   &  MAE         \\
\hline

\multirow{4}*{Weather}   &   96    &       0.166  &   0.220   &   0.169   &   0.223    &    0.175   &   0.230   &   0.171   &   0.224    \\

&   192 &       0.219  &   0.268  &   0.224   &   0.272   &  0.227   &    0.276     &   0.230    &  0.277       \\

&  336   &     0.272   &   0.297  &   0.276   &   0.303    &         0.286   &   0.322     &    0.294   &   0.326    \\

&   720   &       0.355  &    0.360   &   0.362   &   0.368    &          0.366   &    0.379   &    0.384    &      0.387    \\

\hline 

\multirow{4}*{ETTh1}   &   96    &     0.531  &   0.497 
  &   0.538   &   0.501    &   0.543   &   0.506    &   0.557   &   0.519   \\

&   192   &       0.685  &   0.546     &   0.698   &   0.557    &      0.748   &   0.580     &   0.711   &   0.570       \\

&   336   &       0.738  &   0.573   &    0.752   &   0.591    &      0.754   &   0.595     &   0.816    &   0.619    \\

&   720   &       -  &   -  &   -   &   -    &       -   &   -   &   -  &    -         \\

\hline 
\end{tabular}}

\caption{\label{tab:results_few_shot} Results for the few-shot setting. The first 5\% of the training sets used in Table \ref{tab:results_main_long_term} are used for training.  ’-’ means that 5\% time series is not sufficient to constitute a training set.  } 

\end{table*}

\subsection{Ablation studies and analysis}

\noindent \textbf{Ablation on the language model backbones} \quad In our main experiments (Table \ref{tab:results_main_long_term}, \ref{tab:results_main_short_term}, and \ref{tab:results_few_shot}), the LLM we use is LlaMA-2 7B. Thus, it is natural to wonder whether our method works well on the other pre-trained language models. We compare two representative backbones Gemma 2B \cite{banks2024gemma} and GPT-2 large \cite{radford2019language}. The former has 2B parameters, and the latter has 0.5B parameters. The results on the Weather and ETTh1 under the full-data setting (100\% of the training dataset) and few-shot setting (5\% of the training dataset) are reported in Table \ref{tab:results_other_llms}.

\begin{table}[tb!]
\centering
\resizebox{0.48\textwidth}{!}{
\renewcommand\arraystretch{1.2}
\begin{tabular}{cc|cc|cc}
\hline

&   &   \multicolumn{2}{c}{\textbf{Full-data setting}}    &   \multicolumn{2}{c}{\textbf{Few-shot setting}}     \\

\multicolumn{2}{c}{\textbf{Methods}}   &   \textbf{Time-LlaMA}    &     \textbf{Time-LLM}       &   \textbf{Time-LlaMA}    &     \textbf{Time-LLM}       \\ 

\hline

\multicolumn{6}{c}{\emph{Results for Gemma 2B}}   \\

\hline
\multirow{2}*{Weather}   &   96    &   0.153   &  0.157  &    0.169    &   0.173   \\
&  192   &     0.198   &  0.204   &  0.226   &   0.231                 \\

\hline

\multirow{2}*{ETTh1}   &   96    &    0.379   &  0.401   &   0.553   &  0.566     \\
&    192   &     0.421   &  0.432   &   0.706  &  0.718                 \\
\hline 

\multicolumn{6}{c}{\emph{Results for GPT-2 large (0.5B)}}   \\

\hline
\multirow{2}*{Weather}   &   96    &   0.164   &  0.169  &    0.187    &   0.199     \\
&    192   &    0.205  & 0.211   &  0.235  &  0.243                \\
\hline

\multirow{2}*{ETTh1}   &   96    &     0.387   &  0.398   &   0.581   &  0.594    \\
&    192   &     0.432  &   0.438   &   0.727   &   0.742               \\
\hline 
\end{tabular}}

\caption{\label{tab:results_other_llms} Results on the other LLMs. For the few-shot setting, 5\% of the original training set is utilized for training. We report the MSE scores.   } 

\end{table}

With the pre-trained LLM backbone being GPT-2 large or Gemma 2B, the Time-LlaMA method outperforms Time-LLM by clear margins, under both the full-data and few-shot settings, demonstrating the effectiveness of our method with different LLM backbones.

\noindent \textbf{Ablation studies of our Time-LlaMA method} \quad In order to understand the superiority of our Time-LlaMA framework (as in Table Table \ref{tab:results_main_long_term}, \ref{tab:results_main_short_term}, and \ref{tab:results_few_shot}), we now conduct ablation studies on our Time-LlaMA method. We consider the following variants for Time-LlaMA: (a) Time-LlaMA-1, which removes the modality alignment module (Eq \ref{eq:modality_align}). Thus in Time-LlaMA-1, the time-series tokens are directly fed to the LLM backbone. (b) Time-LlaMA-2, which concatenate the text prompt to the left of the time-series tokens, serving as prefix. Naturally, this variant will be less efficient than our main model Time-LlaMA, since it expands the input token sequences. (c) Time-LlaMA-3, which substitutes our D-LoRA mechanism to the static LoRA method. (d) Time-LlaMA-4, which discards the LoRA based fine-tuning module on the LLM backbone and keeps the LLM backbone entirely frozen. 

The experiments are presented in Table \ref{tab:results_ablation_study}. From Table \ref{tab:results_ablation_study}, we can observe that: (a) The comparison between Time-LlaMA-1 to Time-LlaMA demonstrates the necessity of the modality alignment module. (b) Time-LlaMA-2 performs closely to Time-LlaMA, demonstrating that with our modality alignment module, the text prompts containing the task information is no longer needed. (c) The comparison among Time-LlaMA-3, Time-LlaMA-4 and Time-LlaMA shows that by fine-tuning the LLM backbone in a parameter-efficient style helps our Time-LlaMA to achieve superior performance. And our D-LoRA module adaptively adjust which LoRA modules are used to conduct inference for the current test sample, achieving stronger prediction capabilities.

\begin{table}[tb!]
\centering
\resizebox{0.34\textwidth}{!}{
\renewcommand\arraystretch{1.2}
\begin{tabular}{c|cc|cc}
\hline

\multirow{2}*{\textbf{Methods}}  &   \multicolumn{2}{c}{\textbf{Weather }}   &  \multicolumn{2}{c}{\textbf{ETTh1 }}      \\ 

&   96  &  192   &   96   &  192   \\

\hline
Time-LlaMA &   0.166   &   0.219    &   0.531   &   0.685    \\
Time-LlaMA-1 &    0.172  &  0.226   &   0. 538   &   0.697   \\
Time-LlaMA-2 &   0.165   &   0.221   &   0.533   &  0.685  \\
Time-LlaMA-3 &    0.173    &  0.227    &   0.537   &   0.696   \\
Time-LlaMA-4 &   0.178   & 0.232     &   0.542   &   0.705   \\

\hline 
\end{tabular}}

\caption{\label{tab:results_ablation_study} Results for the ablation study, in which each method is trained under the few-shot setting and 5\% of the original training set is utilized for training. We report the MSE scores.   } 

\end{table}

\begin{figure}
\centering
\subfigure{%
\includegraphics[width=0.235\textwidth]{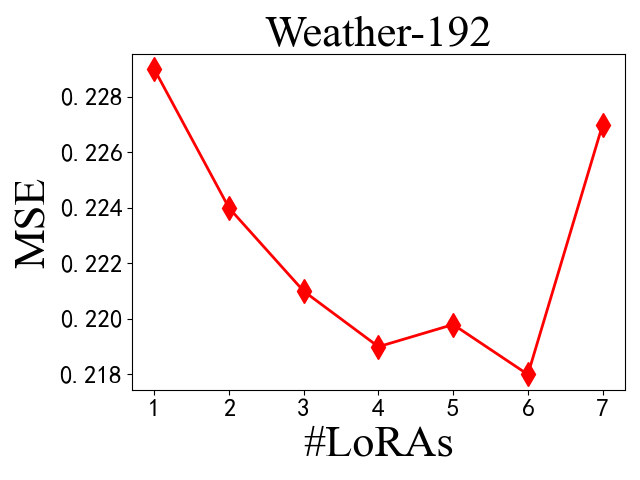}
\label{subfig:Weather-192_num_activated_experts}
}\hspace{-8pt}
\subfigure{%
\includegraphics[width=0.235\textwidth]{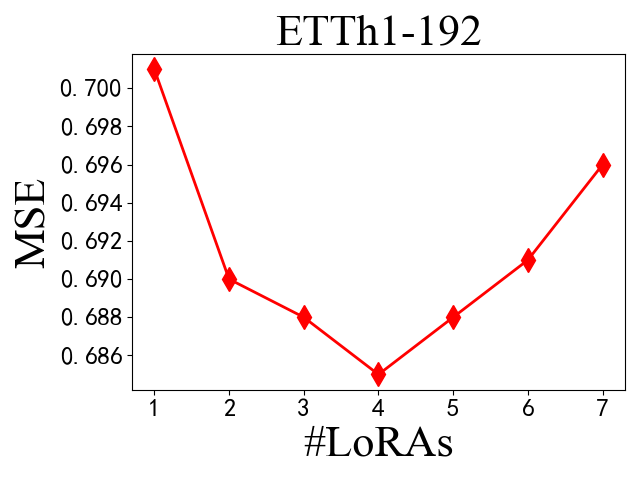}
\label{subfig:ETTh1-192_num_activated_experts}
}
\caption{Performances under different numbers of selected LoRAs per Transformer block. }
\label{fig:different_activated_experts}
\end{figure}

\noindent \textbf{Effects on the number of selected LoRA modules $n$} \quad In our main experiments (Table \ref{tab:results_main_long_term}, ), we set the number of selected LoRA modules $n$ to 4. Now, we alter $n$ to \{1, 2, 3, 5, 6, 7\}, and investigate how this hyper-parameter affects our Time-LlaMA method. The results are demonstrated in Figure \ref{fig:different_activated_experts}. From the experiments, one can see that when $n$ changes from 1 to 7, the performance first becomes better, and then drops. The observations are consistent with ALoRA \cite{liu2024alora}, which demonstrates that reduce the number of LoRA modules per block is beneficial for the LLM's downstream adaptation. 

\noindent \textbf{Efficiency analysis} \quad In our main experiments (Table \ref{tab:results_main_long_term}), we only utilize the first 8 blocks of the LlaMA-2 7B model to encode the time-series information and make predictions. Thus, its inference speed is 5.32 test samples per second on the test set of the Traffic task. Note that in the industrial applications, efficiency is an important factor. Thus, it is of value to compare the latency of our method and the non-LLM method PatchTST. Note that PatchTST transforms the multi-variate time series task like Traffic into multiple single-variate time series tasks. Thus, it has to conduct inference for 862 single-variate series for a single sample in Traffic. Following its original implementations, PatchTST's inference speed is 7.58 samples per second. This comparison demonstrates that through our channel-wise tokenization method, our Time-LlaMA method is actually very efficient, even with LLM backbones.

\begin{figure}[t]
\centering
\includegraphics[width=0.44\textwidth]{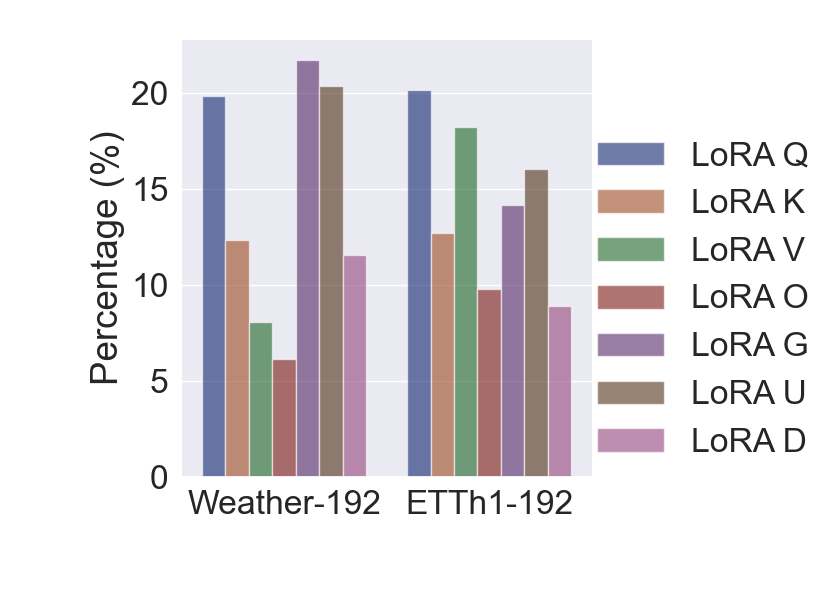} 
\caption{Distribution of activated LoRA experts. }
\label{fig:lora_dist}
\end{figure}

\noindent\textbf{Distributions of the selected LoRAs} \quad We now compare the distribution of LoRA modules across all Transformer layers on the Weather and ETTh1 tasks' test sets (prediction length is 192) in Figure \ref{fig:lora_dist}. We can observe that: (a) different Transformer layers choose to select different LoRA experts via their corresponding routers, and the maximum proportion a LoRA expert can achieve is less than 25\%. The results are intuitive since Transformer layers of different depths represent different knowledge, requiring different LoRA experts to express. (b) the LoRA distributions on different tasks are different. For example, more layers activate LoRA G or LoRA U on the Weather task than on the ETTh1 task.

\section{Conclusion}

In this work, we propose a novel framework, Time-LlaMA. First, Time-LlaMA tokenizes each time series sample by considering each variate as a token. Then we align the time series tokens to the language modality by attending to text prompts' embeddings. Third, the LLM backbone is fine-tuned by a novel LoRA method that adaptively selecting different LoRA modules for different time series samples. Extensive experiments have demonstrated that Time-LlaMA can outperform the recent SOTA baselines.

\section*{Ethical Statement}

In this research, we have carefully considered the ethical implications of developing Time-LlaMA, a framework for time series forecasting using large language models (LLMs). We ensured data privacy by using only publicly available, anonymized, or permitted datasets, avoiding sensitive or proprietary information. To address potential biases, we employed diverse datasets and rigorous testing across domains. We minimized environmental impact by using efficient training techniques like D-LoRA and energy-efficient hardware. Transparency and reproducibility were prioritized through detailed methodology descriptions and plans to release code and model weights. We also acknowledged dual-use concerns, encouraging responsible application of our work, and fostered inclusivity through collaborative and open research practices. These steps align our research with ethical AI development principles.

\section*{Acknowledgments}

None

\bibliographystyle{named}
\bibliography{ijcai25}

\end{document}


\maketitle

\section{Experimental settings}

Now we provide more details for the experiments presented in the main contents. 

\subsection{Implementation}

We mainly follow the experimental configurations in \cite{jin2023time} across all baselines within a unified evaluation pipeline in the Time-Series-Library\footnote{https://github.com/thuml/Time-Series-Library} for fair comparisons. We use Llama-2 7B \cite{Touvron2023Llama2O} as the default backbone model, unless stated otherwise. All our experiments are repeated three times and we report the averaged results. Our method is implemented on PyTorch \cite{paszke2019pytorch} with all experiments conducted on NVIDIA L20 GPUs (48 GB RAM). Our detailed model configurations will be 
, and our code will made available upon acceptance.


\subsection{Datasets}

We evaluate the long-term forecasting (ltf) performance on the well-established eight different benchmarks, including four ETT datasets  (including ETTh1, ETTh2, ETTm1, and ETTm2) from \cite{zhou2021informer}, Weather, Electricity, Traffic, and ILI from \cite{wu2021autoformer}. For short-term time series forecasting (STF), we employ the M4 benchmark \cite{makridakis2018m4}.

\noindent\textbf{ETT} The Electricity Transformer Temperature (ETT) is a crucial indicator in the electric power long-term deployment. This dataset consists of 2 years data from two separated counties in China. To explore the granularity on the Long sequence time-series forecasting (LSTF) problem, different subsets are created, {ETTh1, ETTh2} for 1-hour-level and ETTm1 for 15-minutes-level. Each data point consists of the target value ”oil temperature” and 6 power load features. The train/val/test is 12/4/4 months.

\noindent\textbf{ECL} Measurements of electric power consumption in one household with a one-minute sampling rate over a period of almost 4 years. Different electrical quantities and some sub-metering values are available.This archive contains 2075259 measurements gathered in a house located in Sceaux (7km of Paris, France) between December 2006 and November 2010 (47 months).

\noindent\textbf{Traffic} Traffic is a collection of hourly data from California Department of Transportation, which describes the road occupancy rates measured by different sensors on San Francisco Bay area freeways.

\noindent\textbf{Weather} Weather is recorded every 10 minutes for the 2020 whole year, which contains 21 meteorological indicators, such as air temperature, humidity, etc. 

\noindent\textbf{ILI} The influenza-like illness (ILI) dataset
contains records of patients experiencing severe influenza with complications. 

\noindent\textbf{M4} The M4 benchmark comprises 100K time series, amassed from various domains commonly present in business, financial, and economic forecasting. These time series have been partitioned into six distinctive datasets, each with varying sampling frequencies that range from yearly to hourly. These series are categorized into five different domains: demographic, micro, macro, industry, and finance.

The datasets' statistics are presented in Table \ref{tab:dataset_stats}.

\begin{table*}
\centering
\resizebox{0.92\textwidth}{!}{
\renewcommand\arraystretch{1.1}
\begin{tabular}{@{}ll|c|c|c|c|c}
\hline
\textbf{Tasks} & \textbf{Dataset} & \textbf{Dim.} & \textbf{Series Length} & \textbf{Dataset Size} & \textbf{Frequency} & \textbf{Domain} \\ 
\hline
\multirow{7}{*}{Long-term Forecasting} & ETTm1 & 7 & \{96, 192, 336, 720\} & (34465, 11521, 11521) & 15 min & Temperature \\
& ETTm2 & 7 & \{96, 192, 336, 720\} & (34465, 11521, 11521) & 15 min & Temperature \\
& ETTh1 & 7 & \{96, 192, 336, 720\} & (8545, 2881, 2881) & 1 hour & Temperature \\
& ETTh2 & 7 & \{96, 192, 336, 720\} & (8545, 2881, 2881) & 1 hour & Temperature \\
& Electricity & 321 & \{96, 192, 336, 720\} & (18317, 2633, 5261) & 1 hour & Electricity \\
& Traffic & 862 & \{96, 192, 336, 720\} & (12185, 1757, 3509) & 1 hour & Transportation \\
& Weather & 21 & \{96, 192, 336, 720\} & (36792, 5271, 10540) & 10 min & Weather \\
& ILI & 7 & \{24, 36, 48, 60\} & (617, 74, 170) & 1 week & Illness \\ \hline

\multirow{6}{*}{Short-term Forecasting} & M4-Yearly & 1 & 6 & (23000, 0, 23000) & Yearly & Demographic \\
& M4-Quarterly & 1 & 8 & (24000, 0, 24000) & Quarterly & Finance \\
& M4-Monthly & 1 & 18 & (48000, 0, 48000) & Monthly & Industry \\
& M4-Weakly & 1 & 13 & (359, 0, 359) & Weakly & Macro \\
& M4-Daily & 1 & 14 & (4227, 0, 4227) & Daily & Micro \\
& M4-Hourly & 1 & 48 & (414, 0, 414) & Hourly & Other \\ 
\hline

\end{tabular}}
\caption{Dataset statistics. The dimension indicates the number of time series (i.e., channels), and the dataset size is organized in (training, validation, testing).}
\label{tab:dataset_stats}
\end{table*}

\subsection{Evaulation metrics}

We now specify the evaluation metrics we used for comparing different models. We utilize the mean square error (MSE) and mean absolute error (MAE) for long-term forecasting. For the short-term forecasting task on M4 benchmark, we adopt the symmetric mean absolute percentage error (SMAPE), mean absolute scaled error (MASE), and overall weighted average (OWA), following \cite{oreshkin2019n}. The calculations of these metrics are as follows:
\begin{align}
\text{MSE} &= \frac{1}{H} \sum_{h=1}^{T} (\mathbf{Y}_h - \hat{\mathbf{Y}}_h)^2,  \\
\text{MAE} &= \frac{1}{H} \sum_{h=1}^{H} |\mathbf{Y}_h - \hat{\mathbf{Y}}_h|,  \\
\text{SMAPE} &= \frac{200}{H} \sum_{h=1}^{H} \frac{|\mathbf{Y}_h - \hat{\mathbf{Y}}_h|}{|\mathbf{Y}_h| + |\hat{\mathbf{Y}}_h|},  \\
\text{MAPE} &= \frac{100}{H} \sum_{h=1}^{H} \frac{|\mathbf{Y}_h - \hat{\mathbf{Y}}_h|}{|\mathbf{Y}_h|},  \\
\text{MASE} &= \frac{1}{H} \sum_{h=1}^{H} \frac{|\mathbf{Y}_h - \hat{\mathbf{Y}}_h|}{\frac{1}{H-s} \sum_{j=s+1}^{H} |\mathbf{Y}_j - \mathbf{Y}_{j-s}|},  \\
\text{OWA} &= \frac{1}{2} \left[ \frac{\text{SMAPE}}{\text{SMAPE}_{\text{Naive}}} + \frac{\text{MASE}}{\text{MASE}_{\text{Naive}}} \right],  \\
\end{align}
where $ s $ is the periodicity of the time series data. $ H $ denotes the number of data points (i.e., prediction horizon in our cases). $ \mathbf{Y}_h $ and $ \hat{\mathbf{Y}}_h $ are the $ h $-th ground truth and prediction where $ h \in \{1, \cdots, H\} $.

\subsection{Configurations for training}

The Adam optimizer \cite{losh

chilov2017decoupled} is employed throughout all experiments. The loss objective is MSE for the long-term forecasting tasks, and SMAPE for the short-term forecasting tasks. The learning rate is denoted as $\text{LR}$. We utilize the LlaMA-2 7B \cite{Touvron2023Llama2O} model, maintaining the backbone model layers at 8 across all tasks. Denote the lookback window's length as $ T_L $, the prediction horizon as $T_P$. And the heads $ K $ correlate to the multi-head cross-attention utilized for time-series data reprogramming. For the LoRA modules, the number of ranks $r$ is set to $8$. Each Transformer block's LoRA router activates $n = 4$ LoRA modules. We detail the configurations for each task in Table \ref{tab:configurations}.

\begin{table*}
\centering
\resizebox{0.98\textwidth}{!}{
\renewcommand\arraystretch{1.1}
\begin{tabular}{l|c|c|c|c|c|c|c|c|c|c}
\toprule
\multirow{2}{*}{\textbf{Task-Dataset }} & \multicolumn{6}{c|}{\textbf{Model Hyperparameter}} & \multicolumn{4}{c}{\textbf{Training Process}} \\
\cmidrule(lr){2-7} \cmidrule(lr){8-11}
& \textbf{ Layers} & \textbf{ $ T_{L} $} & \textbf{ $T_P$} &   \textbf{ $K$ }      &     \textbf{ $r$ }    &  \textbf{ $n$ }    &     \textbf{LR*} & \textbf{Loss} & \textbf{Batch Size} & \textbf{Epochs} \\
\midrule
LTF - ETTh1 &  8 & 512 & \{96, 192, 336, 720\} & 8 &  8  &   4  &  $ 10^{-3} $ & MSE &  16  & 20 \\
LTF - ETTm1 &  8 & 512 & \{96, 192, 336, 720\} & 8 &  8  &    4 & $ 10^{-3} $ & MSE & 16 & 20 \\
LTF - Weather &  8 & 512 & \{96, 192, 336, 720\} & 8 & 8  &    4 & $ 10^{-3} $ & MSE & 16 & 20 \\
LTF - Electricity &  8 & 512 & \{96, 192, 336, 720\} & 8 &  8  &   4 & $ 10^{-2} $ & MSE & 16 & 20 \\
LTF - Traffic &  8 & 512 & \{96, 192, 336, 720\} & 8 & 8  &    4 & $ 10^{-2} $ & MSE & 12 & 20  \\
LTF - ILI &  8 & 96 & \{24, 36, 48, 60\} & 8 & 8  &   4 & $ 10^{-2} $ & MSE & 16 & 20 \\
STF - M4 &  8 & $ 2 \times T_{P} $ &  \{6, 48\}  & 8   & 8  &    4  & $ 10^{-3} $ & SMAPE & 32 & 30 \\

\bottomrule

\end{tabular}}
\caption{An overview of the experimental configurations for TIME-LlaMA. LTF and STF denote long-term and short-term forecasting, respectively. }
\label{tab:configurations}
\end{table*}

\bibliographystyle{named}
\bibliography{ijcai25}